\documentclass[10pt,twocolumn,letterpaper]{article}

\usepackage{iccv}
\usepackage{times}
\usepackage{epsfig}
\usepackage{graphicx}
\usepackage{amsmath}
\usepackage{amssymb}

\usepackage{multicol} % 多栏布局宏包
\usepackage{diagbox} % 对角线包
\usepackage{makecell}
\usepackage{multirow} % 多行布局

\usepackage [font=normalsize,labelfont=normalsize] {caption}

\usepackage[ruled,vlined]{algorithm2e}

\usepackage{url}

\usepackage{enumitem}

\usepackage{hyperref}
\hypersetup{hypertex=true,
colorlinks=true,
linkcolor=red,
anchorcolor=blue,
citecolor=black}

\usepackage{tikz,pgfplots}					% 包导入

\usepackage{graphicx} % 需使用包
\usepackage{float}
\usepackage{subfigure}

\usepackage{colortbl}

\usepackage{amssymb}
\usepackage{amsmath}
\usepackage{bbm}
\usepackage{marvosym} % letter symbol

\newcommand{\yao}[1]{{\textcolor{orange}{yao: #1}}}

\usepackage[]{hyperref}         % ...clickable refs within pdf...
\hypersetup{                    % ...like so
colorlinks,
linkcolor={green!80!black},
citecolor={red!70!black},
urlcolor={blue!70!black}
}
\iccvfinalcopy % *** Uncomment this line for the final submission

\def\iccvPaperID{****} % *** Enter the ICCV Paper ID here

\ificcvfinal\fi

\begin{document}

%%%%%%%%% TITLE
\title{ Dual-level Interaction for Domain Adaptive Semantic Segmentation}

\iffalse
\author{Dongyu Yao Boheng Li\textsuperscript{\Letter}\\
School of Cyber Science and Engineering, Wuhan University\\
Institution1 address\\
{\tt\small \{dongyu.yao, antigonerandy\}@whu.edu.cn}
}
% \name{Dongyu Yao, Boheng Li\textsuperscript{\Letter} 
\thanks{\textsuperscript{\Letter}Corresponding author}
%Address and e-mail should NOT be added in the submission paper. They should be present only in the camera ready paper. 
% competition version
%\address{School of Cyber Science and Engineering, Wuhan University \\
%\{dongyu.yao, antigonerandy\}@whu.edu.cn
%}
\fi

% \author{Dongyu Yao, 
%         Boheng Li$^*$\\
% School of Cyber Science and Engineering, Wuhan University, China\\
% % Institution1 address\\
% {\tt\small \{dongyu.yao, boheng.li\}@whu.edu.cn}\thanks{Boheng Li is the corresponding author.}}
% For a paper whose authors are all at the same institution,
% omit the following lines up until the closing ``}''.
% Additional authors and addresses can be added with ``\and'',
% just like the second author.
% To save space, use either the email address or home page, not both

\author{Dongyu Yao, Boheng Li\thanks{ Corresponding author}\\
        School of Cyber Science and Engineering, Wuhan University, China \\
	\small {\tt\{dongyu.yao, boheng.li\}@whu.edu.cn} }
\iffalse
\author{Anonymous Authorsr\\
Institution1\\
Institution1 address\\
{\tt\small firstauthor@i1.org}
% For a paper whose authors are all at the same institution,
% omit the following lines up until the closing ``}''.
% Additional authors and addresses can be added with ``\and'',
% just like the second author.
% To save space, use either the email address or home page, not both
\and
Second Author\\
Institution2\\
First line of institution2 address\\
{\tt\small secondauthor@i2.org}
}
\fi

\maketitle
% Remove page # from the first page of camera-ready.
\ificcvfinal\thispagestyle{empty}\fi

%%%%%%%%% ABSTRACT
\begin{abstract}
Self-training approach recently secures its position in domain adaptive semantic segmentation, where a model is trained with target domain pseudo-labels. 
Current advances have mitigated noisy pseudo-labels resulting from the domain gap.
However, they still struggle with erroneous pseudo-labels near the boundaries of the semantic classifier. %, which . 
In this paper, we tackle this issue by proposing a dual-level interaction for domain adaptation (DIDA) in semantic segmentation. Explicitly, we encourage the different augmented views of the same pixel to have not only similar class prediction (semantic-level) but also akin similarity relationship with respect to other pixels (instance-level). 
As it's impossible to keep features of all pixel instances for a dataset, %we therefore introduce a instance memory bank for source domain data to selectively store the informative features of pixels as instances.
we, therefore, maintain a labeled instance bank with dynamic updating strategies to selectively store the informative features of instances.
Further, DIDA performs cross-level interaction with scattering and gathering techniques to regenerate more reliable pseudo-labels. Our method outperforms the state-of-the-art by a notable margin, especially on confusing and long-tailed classes. Code is available at \href{https://github.com/RainJamesY/DIDA}{https://github.com/RainJamesY/DIDA} 
% We will opensource our code upon publication.
\end{abstract}

%-----------------OUR DIDA-------------------------

\section{Introduction}
\label{sec:intro}

% \Wang{The writing should be problem-driven, not method-driven.}

Semantic segmentation, aiming at assigning a label for every single pixel in a given image, is a fundamental task in computer vision.%  and high-level image understanding. 
% While deep learning methods \cite{Deeplab, overconfident} have become the go-to choice in this field, most of them require tremendous quantities of annotated data, which is both time and labor consuming, for high performance. Thus, researchers suggest to use realistic synthetic data (\eg{}, from virtual simulation \cite{synthia} or open-world games \cite{gta}) with automatically generated pixel-level label annotations for training augmentation.
Learning with synthetic data (\eg{}, from virtual simulation \cite{synthia} or open-world games \cite{gta}) has revolutionized segmentation tasks over the past few years, which effectively saves time and labor from the onerous pixel-level annotations.
However, the existence of domain shifts between the rendered synthetic data and real-world distributions severely reduces the models' performance \cite{ProDA}. To mitigate this problem, Unsupervised Domain Adaptation (UDA) is explored  to generalize the network trained with labeled source (synthetic) data to unlabeled target (real) data. 
% As training with synthetic data becomes the recent trend, coping with such domain shifts is currently a key challenge to the community. 
% \Wang{The issues of applying UDA in the community. or the inevitable weakness of the existing studies.}

\begin{figure}[t]
\centering  %图片全局居中
\includegraphics[scale=0.3]{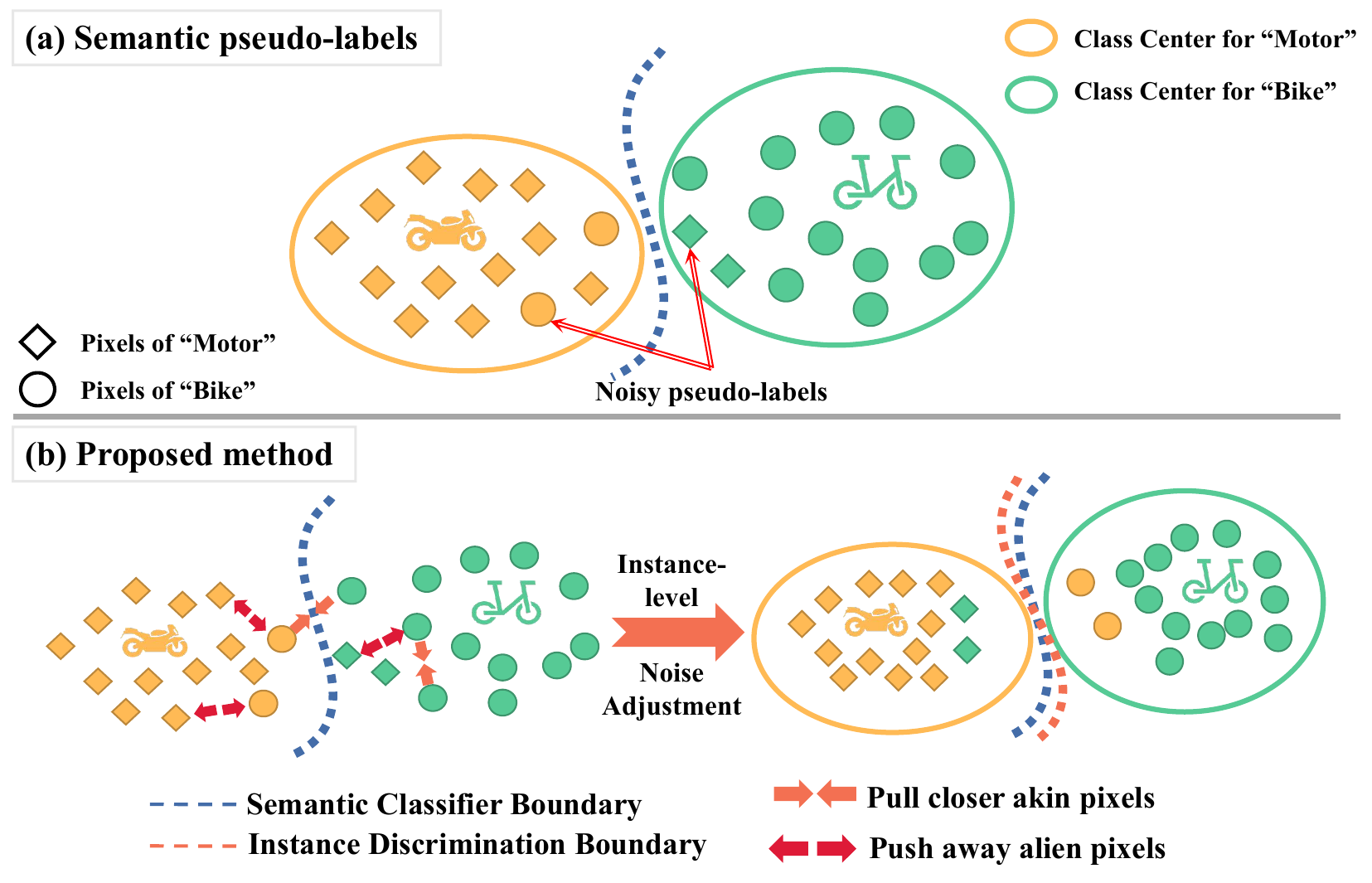}
\caption{\small{\textbf{Intuition behind DIDA.} \textbf{(a)} The semantic classifier trained on the source domain can be viewed as class feature centers, which possess a natural weakness in classifying pixels near or across the category boundaries thus producing erroneous and noisy semantic pseudo-labels. \textbf{(b)} In DIDA, we utilize instance-level discrimination with proposed instance loss to adjust noisy pseudo-labels. By simultaneously considering the semantic-level and instance-level information with cross-level interaction, we reset the classification boundaries for more robust pseudo-labeling.}}
\label{intro_fig}
\end{figure}

% \Wang{Clarify the problem, the efforts of existing studies in tackling this problem, present the shortcomings of existing studies. The challenges of addressing this problem.}
\noindent\textbf{Existing Work:} 
% There is a rich line of works on UDA, which can be broadly categorized into adversarial training and self-training. Herein, we focus on self-training approaches because adversarial-based methods require auxiliary models, which are less practical in real scenarios.
% A substantial body of literature exists on UDA, with a predominant focus on a methodology known as self-training. 
% Self-training entails initially training the network using labeled synthetic data. Subsequently, the pre-trained network is employed to generate pseudo-labels for real data, where finally these pseudo-labeled real data are utilized to re-train the model.
Most of the existing works on self-training UDA \cite{overconfident, DACS, ProDA} generate target domain pseudo-labels based on the semantic-level class predictions of the network for future self-training.
% but also currently underperforms the latter. 
% Most of the existing works on self-training UDA \cite{overconfident, DACS, ProDA} generate target domain pseudo-labels based on the semantic-level class predictions of the network for future self-training. 
% For example, Zou \etal~\cite{overconfident} suggests setting a confidence threshold to select more reliable semantic-level pseudo-labels, improvements like DACS \cite{DACS} which introduces consistency regularization techniques and ProDA \cite{ProDA} that relies on the soft prototypical assignment were proposed to generate more accurate and informative ones.
To step further, a very recent work DAFormer \cite{DAFormer} utilizes a Transformer encoder and a multilevel feature fusion decoder architecture, and designed three training strategies to stabilize training as well as avoid overfitting, which achieved state-of-the-art performance.
% However, one problem of the existing works is that the semantic-level pseudo-labels inevitably contain noises, which would cause the well-known ``overconfidence" issue \cite{overconfident} if without proper noise adjustment. Though previous methods \cite{ProDA} proposed vigorous strategies towards solving this problem, it was still challenging to clearly sort out noisy pseudo-labels close to the classification boundary as shown in  Figure \ref{intro_fig}\textcolor{red}{(a)}. Furthermore, the poor adaptation performance on long-tailed and analogous categories also urgently requires a solution. 
% Unfortunately, due to the natural weakness (Figure \ref{intro_fig}\textcolor{red}{(a)}) that the semantic classifier possesses,
% these methods still struggle with clearly sorting out noisy pseudo-labels close to the classification boundary.
% Furthermore, the ``overconfidence" issue \cite{overconfident}
% caused by long-tailed, analogous and overlapping  categories also urgently requires a solution.  

One problem of the existing methods is that they still preserve a large number of noisy pseudo-labels, especially those near the decision boundaries (Figure \ref{intro_fig}\textcolor{red}{(a)}). This is because they  only obtain the pseudo-labels through the semantic classifier and eliminate the unreliable pseudo-labels via selecting a confidence threshold, which is rather an empirical process varying across tasks, models, and datasets \cite{CBST}. It is extremely difficult to formulate such selection as a mathematical function, making it hard to optimize. Furthermore, the presence of confusing (\eg{}, analogous and adjacent/overlapping) categories leads to an open issue known as ``overconfidence'' \cite{overconfident}, which significantly degrades the performance on these categories (Table \ref{comparison} and \ref{comparison1}).

%Semantic segmentation is a fundamental computer vision task that requires models to assign semantic labels to all pixels in a given image. 
%While deep learning methods \cite{Deeplab, SegNeXt} have achieved great success in this field, most of them are heavily dependent on tremendous quantities of labeled data to attain high performance. Nevertheless, annotating such pixel-wise images is time and labor consuming, which drives people to circumvent this issue by utilizing more easy-to-get synthetic images \cite{gta, synthia}. Moreover, the existence of domain shifts between synthetic data (labeled source domain) and real-world distributions (unlabeled target domain) 
%severely reduces the models' performance \cite{ProDA}. To heal the domain gap, unsupervised domain adaptation (UDA) has been explored.

% classify existing methods; 
% One technique originally proposed in semi-supervised learning (SSL) for classification is self-training (or pseudo-labelling), which generates target domain pseudo-labels based on the class predictions of the network. The idea was later well adapted to UDA for segmentation \cite{DACS, ProDA} and a very recent work DAFormer \cite{DAFormer} became the state-of-the-art.
% However, these approaches only focus on the semantic-level discrimination and their adaptation performance is in most cases restricted to the semantic pseudo-labels that would contain noise without proper adjustment, resulting in the ``overconfidence" issue \cite{overconfident} towards confident but erroneous predictions.

\noindent\textbf{Our Work:} To alleviate the aforementioned limitations, in this paper, we propose to leverage \textbf{\underline{D}}ual-level \textbf{\underline{I}}nteraction for \textbf{\underline{D}}omain \textbf{\underline{A}}daptive (DIDA) semantic segmentation.  Inspired by previous work~\cite{saito2020universal} that the fine-grained instance-level discrimination could be conducive to adjusting noisy semantic-level pseudo-labels, especially those around classifier boundaries, in DIDA, we simultaneously consider the semantic-level and instance-level consistency regularization. 
% fig 1b, detailed adjusting process
Specifically, we encourage different (weakly and strongly) augmented views of the same pixel to have not only consistent semantic class prediction but also akin similarity relationship towards other pixels (instance-level discrimination) to provide additional instance-feature consistency beyond the semantics.
% After discriminating the more intrinsic similarity among instance-level pixels, 
As a result, we adjust the distribution of semantic pseudo-labels with proposed instance loss and reset the classification boundaries
(Figure \ref{intro_fig}\textcolor{red}{(b)}). 

In the semantic segmentation task, the main challenge of introducing instance-level discrimination is that features of pixel instances (instead of image-wise instances in classification tasks % \cite{ICCV_classification}
) can cause tremendous extra storage (\eg{}, 50 billion instances for the GTA5 dataset). 
% \yao{As the instances in semantic segmentation are pixels rather than images, it's impossible to keep track of features of all pixels for a dataset (\eg{}, 50 billion instances for GTA5 dataset). We therefore introduce a instance memory bank for source domain data to selectively store the informative features of pixels as instances.}
%On large, high-resolution datasets, the storage consumption of instance-level features even goes beyond the limit of any existing hardware. 
To overcome this issue, we design a labeled \emph{Instance Bank (IB)} to selectively deposit instance features rather than keeping the entirety. We dynamically update our IB via our class-balanced sampling (CBS) and boundary pixel selecting (BPS) strategies. By doing so, DIDA enriches the fine-grained instance-level pseudo-labels with long-tailed and analogous categories for future self-training, and thus the model becomes more at ease in dealing with such tricky categories.
%Moreover, we introduce a novel instance loss to encourage instance characteristic predictions from target domain samples to align with those deposited in the IB. 
Different from previous methods that naively calculate on semantic pseudo-labels \cite{DACS, ProDA, DAFormer} or incorporate instance information by simply adding loss components \cite{liu2021domain, BAPA-Net, cluda}, we further present \emph{scattering} and \emph{gathering} techniques to interact predictions from both levels, thus generating less noisy training targets. 

Considering the strong similarity among fine-grained computer vision tasks, % (\eg{}, object detection, instance segmentation),
our framework shows prominent applicability to other self-training scenarios.

Our key contributions are summarized as follows:
\begin{itemize}[itemsep=1.4pt,topsep=0pt,parsep=0pt] 
% [itemsep=2pt,topsep=0pt,parsep=0pt]
  \item We propose DIDA, a novel framework that exploits both semantic-level and instance-level consistency regularization for better noise adjustment. To the best of our knowledge, in the task of UDA for semantic segmentation, we are the first attempt that allows the information (the pseudo-labels) of two levels to calibrate and interact with each other.
  % \Wang{To the best of our knowledge, we are the first attempt to consider instance-level information in the task of UDA in semantic segmentation. TODO Wangrun}
  \item We design a labeled instance bank to overcome the issue of storage in incorporating the instance-level information and devise class-balanced sampling and boundary pixel selecting strategies to enhance the performance on long-tailed and analogous categories. % We present interactive techniques to align both predictions and regenerate more reliable pseudo-labels.
  % We also introduce an instance loss and perform interaction of both levels with \emph{scattering} and \emph{gathering} techniques. \Wang{Highlight just a key point for implementation. show the results.}
  % To cope with the large storage of pixel instance features, we innovatively  maintain a labeled feature bank with dynamic updating strategies to filter out useless ones.
  \item The proposed DIDA notably outperforms the previous state-of-the-art. % methods by a notable margin. %, as shown in Figure \ref{progress_figure}.
  %Especially on the confusing classes such as ``sidewalk" and ``truck" as well as long-tailed classes such as ``train" and ``sign". especially on the confusing and long-tailed classes.
  On GTA5 $\rightarrow$ Cityscapes adaptation, we improve the mIoU from 68.3 to 71.0 and on SYNTHIA $\rightarrow$ Cityscapes from 60.9 to 63.3. 
  Especially, DIDA shows outstanding IoU results  on the confusing classes such as ``sidewalk" and ``truck" as well as long-tailed classes such as ``train" and ``sign". 
  \iffalse
  Especially, DIDA shows outstanding IoU results on confusing categories such as ``sidewalk" from 70.2 to 78.0,
  % and ``truck" from 74.5 to 80.9, and ``sign" from 59.4 to 66.1 
  and long-tailed categories such as ``train" from 65.1 to 73.8 on GTA5$\rightarrow$Cityscapes.
  % Extensive ablation studies also demonstrated the effectiveness of our strategies. \Wang{The experimental results should be simplified, just give few numbers.}
  \fi
\end{itemize}

\section{Related Work}

\noindent
\textbf{Domain Adaptation for semantic segmentation.} 
%Unlike UDA for image recognition tasks \cite{tcsvt-UDA1, } that rely on small and simple image-wise datasets, the segmentation task is higher-structured far more challenging since every single pixel needs to be classified. 
% To circumvent the expensive annotations of real-world images, UDA for semantic segmentation emerged as a hot topic. 
Multiple methods have been proposed to bridge the domain gap between synthetic data and real ones. 
Compared to adversarial training methods that align distributions of source and target domain at different levels, \ie, input-level \cite{ad-input1, cycada}, feature-level \cite{feature, ad-feature4}, and output-level \cite{advent, ad-output1}, 
the self-training approaches obtain more competitive results. By generating target domain pseudo-labels and iteratively refining (self-training) the model using the most confident ones \cite{SAC, caco, self-train1, CBST}, the model's performance is further improved. 
However, the naive generation of target domain pseudo-labels is error-prone, causing the network to converge in the wrong direction. Therefore, several works were proposed to ``rectify" the erroneous pseudo-labels on the semantic-level \cite{conf-measure1, conf-measure2, pre-train1, densification1}.
% with techniques like output confidence measurement \cite{conf-measure1, conf-measure2}, the introduction of pre-training phase \cite{pre-train1}, as well as densification frameworks \cite{densification1}. 
Following this trend, 
DACS \cite{DACS} used data-augmented consistency regularization \cite{mixmatch} to mix source and target images during training. 
ProDA \cite{ProDA} employed correction of pseudo-labels with feature distances to prototypes. 
% add more citations
Lukas \etal \cite{DAFormer} referred to the UDA strategy from DACS \cite{DACS} and demonstrated state-of-the-art performance using Transformer as the backbone.
There also exist works that exploit instance information, for example, Liu \etal devised a patch-wise contrastive learning framework, BAPA-Net \cite{BAPA-Net} encouraged prototype alignment at the class-level, CLUDA \cite{cluda} incorporated contrastive loss using target domain semantic pseudo-labels.
However, the above methods either solely focused on semantic pseudo-label rectification or naively incorporated instance information by additional loss components.
% All these methods above, nevertheless, solely refine and calculate pseudo-labels of the semantic-level (the pseudo-labels are all obtained by the semantic classifier), while neglecting the internal pixel-wise structures of the instance-level. 
In contrast, our method introduces instance consistency regularization as an auxiliary classifier to produce pseudo-labels with different noisy patterns and further adjusts pseudo-labels of the two levels with interactive techniques.

\vspace{1\baselineskip}

% Additional related work
\noindent
\textbf{Consistency Regularization.}
Consistency Regularization is first explored in Semi-supervised Learning (SSL) \cite{mixmatch, fixmatch} and is recently adapted to Unsupervised Domain Adaptation (UDA) \cite{cutmix, classmix, UDA-seg1, UDA-seg2}. 
The core idea of Consistency Regularization is to encourage the model to produce similar output predictions for the same input/feature with different perturbations, \eg{}, the input perturbation methods \cite{perturbation1, cutmix} randomly augment the input images with different augmentation degree and the feature perturbation \cite{feature-pert1, feature-pert2} is generally applied by using multiple decoders and supervising the consistency between the outputs of different decoders. Since our dual-level interaction method mainly explores the intrinsic pixel structures of an image, we optimize our model from a batch of differently augmented target input. Moreover, recent advances \cite{CE1, DACS, classmix} concluded that the Cross-Entropy (CE) loss performs better in the fine-grained segmentation task than the Mean Square Error (MSE) loss and Kullback-Leibler (KL) loss. Therefore, in this work, we adopt the CE loss to enforce the consistency of instance-level discrimination.
% \yao{TODO: Write about our semantic training framework -- based on mean teacher, for consistency regularization; CE-loss is superior; instance-level consistency is performed with CE-loss}

\section{Method}
\begin{figure*}[t]
\centering
	\includegraphics[width=1.0\linewidth]{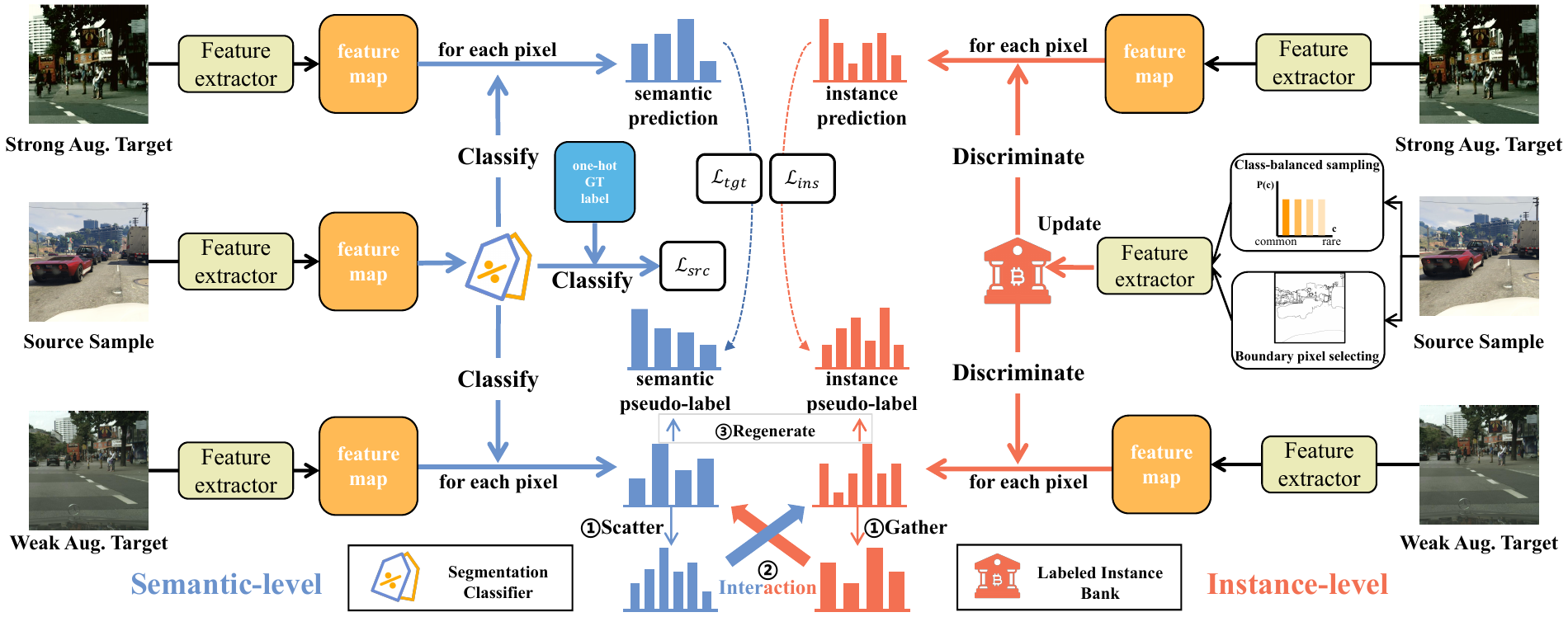}
\caption{\small Overview of DIDA framework. 
The \textbf{black stream} of data augmentation and feature extraction process produces feature maps from an identical batch of input images. The \textbf{blue} and \textbf{red} streams are the semantic-level
and instance-level self-training process, respectively,
which can be viewed as consistency regularization on
both levels produce the $\mathcal{L}_{tgt}$ and $\mathcal{L}_{ins}$.
DIDA adopts class-balanced sampling (CBS) and boundary pixel selecting (BPS) to filter useful source instance features and deposit them into the labeled instance bank (IB) (Sec. \ref{update-bank}). We use the source instance features (stored in IB) as the ``instance classifier" and discriminate between source and target instances to produce instance pseudo-labels (Sec. \ref{discrimination}). The pseudo-label regeneration (bottom middle) calibrates pseudo-labels from the 
two levels (with different channel sizes), then regenerates less noisy pseudo-labels to replace the naive ones (Sec. \ref{interaction}).
During each iteration, the source domain training process 
of $\mathcal{L}_{src}$ and the self-training process of two levels start simultaneously and obtain the summed overall training 
objective $\mathcal{L}_{overall}$.}
\label{overview}
\end{figure*}
% \vspace{-15pt}

\subsection{Preliminaries}
\label{preliminaries}
In this section, we introduce the preliminary semantic-level self-training method for UDA \cite{CBST, DACS, DAFormer}. Given the source domain images $\mathcal{X}_{s}=\left\{x_{s}\right\}_{j=1}^{n_{s}}$  with segmentation labels $\mathcal{Y}_{s}=\left\{y_{s}\right\}_{j=1}^{n_{s}}$, a neural network is trained to obtain useful knowledge from the
source and is expected to achieve good performance on the target images $\mathcal{X}_{t}=\left\{x_{t}\right\}_{j=1}^{n_{t}}$ without the target ground truth labels. In the following sections, we use $i$ to note the $i$-th pixel in an image and $j$ to note the $j$-th image sample in datasets from each domain.

In a typical UDA pipeline, we first train a neural network $h$ with labeled source data. The network $h$ can be divided into $h = f \circ g$, where $f(\cdot)$ is a feature extractor that extracts a feature map $\mathbf{m}=f(x_{s})$ from a given source image $x_{s}$, and $g(\cdot)$ is the fully connected pixel level classifier which is employed to map $\mathbf{m}$ into semantic predictions, written as $p_s = g(\mathbf{m})$. Afterward, the source domain samples are directly optimized with a categorical cross-entropy (CE) loss:
{\small
\begin{eqnarray}
\mathcal{L}_{src}^{}=-\sum_{i=1}^{H \times W} \sum_{c=1}^{C} y_{s}^{(i, c)} \log (p_s^{(i, c)}),
\end{eqnarray}}
where $p_s^{(i, c)}$ represents the softmax probability of pixel $x_s^{(i)}$ belonging to the $c$-th class. A similar definition applies to $p_t^{(i, c)}$. Since the naive network $h=f \circ g$ trained with source data does not generalize well to the target domain owing to the domain gap, the self-training approaches first assign semantic pseudo-labels to the images from the target domain and then train $h$ with the pseudo-labeled images. A conventional method is to use the most probable class predicted by $h$ as the semantic pseudo-labels:
% \iffalse
% In the online generation \cite{AugCon, ContextMix, ProDA} process, a teacher network $h_\phi$ whose weights are updated as the exponential moving average of weights in $h$ after each training step $u$ \cite{ema}
% \begin{eqnarray}
% \phi_{u+1} \leftarrow \alpha \phi_{u}+(1-\alpha) h_{u}
% \end{eqnarray}
% is maintained to stably generate semantic pseudo-labels $\hat{y}_{t}$:
% \fi
{small
\begin{eqnarray}
\hat{y}_{t}^{(i, c)}=\left\{\begin{array}{ll}
1, & \textrm{if} \quad  c=\underset{c^{\prime}}{\arg \max } \, p_t^{\left(i,  c^{\prime}\right)}, \\
0, & \textrm{otherwise}
\end{array}\right.
\end{eqnarray}
}
Evidently, this ``brute force'' strategy suffers from the noisy-label problem \cite{DACS, ProDA}. This is because the pixels near the decision boundary are likely to be assigned with wrong pseudo-labels. A typical method to mitigate this problem is to set a confidence threshold $\tau$ to filter target sample pixels whose largest class probabilities in the pseudo-labels are larger than $\tau$ \cite{DACS,DAFormer}.
\iffalse
\begin{eqnarray}
q_{t}^{(i)}=\frac{\sum_{i=1}^{H \times W}\left[\max _{c^{\prime}} p_t^{\left(i,  c^{\prime}\right)}>\tau\right]}{H \cdot W} .
\end{eqnarray}
\fi
In this way, the unsupervised semantic-level classification loss on the target domain can be defined as:
{\small
\begin{eqnarray}
\label{semantic-pseudo}
\mathcal{L}_{tgt}^{}=-\sum_{i=1}^{H \times W} \sum_{c=1}^{C} \mathbbm{1}\left(\max \, \hat{y}_{t}^{(i, c)}>\tau\right)  \log (p_t^{(i, c)}).%q_{t}^{(i)} 
\end{eqnarray}
}
Unfortunately, an open problem of the existing threshold-based methods is that it is extremely difficult to find a ``perfect'' threshold that can exclude all noisy labels. This is because the selection of threshold is rather an empirical process \cite{CBST}, which varies between tasks and datasets. Thus it is challenging to formulate it as a general mathematical function, making this problem hard to optimize. In this paper, we address this problem from a totally different viewpoint: seeking the regularization of the semantic pseudo-labels from the instance-level perspective.

\subsection{Instance-Level Discrimination }
\label{discrimination}
%This part details our proposed approach of considering instance-level discrimination as previously stated. 
For this component, we view each pixel instance
as a distinct class of its own and adjust our model to distinguish between ``individual" instance classes. 

Firstly, we generate the source domain image-wise feature map $\mathbf{m}_s\in \mathbf{R}^{[H\times W] \times D}$ ($D$ is the channel size of the extracted feature). Noticeably, $\mathbf{m}_s$ consists of $H\times W$ corresponding pixels, and each pixel possesses an embedding $e_s$ with shape $[1 \times 1] \times D$, $e_s \in \mathbf{m}_s$. As mentioned earlier, it is impossible to store all extracted pixel-level features in the memory, thus we selectively store some of them in the memory bank $\mathcal{B}$ (see updating strategy in Sec. \ref{update-bank}). We denote the embeddings of pixels deposited in the bank $ \mathcal{B}$ (obtained from source domain image) as $\left\{e_{k}: k \in(1, \ldots, K) \right\}$. 
Likewise, $\mathbf{m}^A_t$ and $\mathbf{m}_t^\alpha$ are feature maps obtained from strongly-augmented and weakly augmented target samples, where $e^A_t \in \mathbf{m}^A_t$ and $e_t^\alpha \in \mathbf{m}_t^\alpha$ are used to represent instance embeddings.
The source domain features in the bank serve as our auxiliary ``instance classifier" that goes beyond the limitation of the semantic decision boundary.

Under the non-parametric softmax formulation \cite{instance-dis}, for strongly augmented target pixel instance $x_t^{A(i)}$ with embedding $e^A_t$, we use the $cosine$ $similarity$  $\mathbf{cos\_sim}(\mathbf{u}, \mathbf{v})=\mathbf{u}^{T} \mathbf{v} /\|\mathbf{u}\|\|\mathbf{v}\|$ to calculate its instance-level similarity with each sample in $ \mathcal{B}$, as the instance-level prediction of $x_t^{A(i)}$:
{\small
\begin{equation}
\begin{aligned}
 \hat{y}_{ins}^{(i)} &= [q_{1}^A, \dots,q_{k^\prime}^A, \dots] \\ \textrm{where} &\quad q_{k^\prime}^A =\frac{\exp \left((e_t^A)^{T} e_{k^\prime} / tp\right)}{\sum_{k=1}^{K} \exp \left((e_t^A)^{T} e_{k} / tp\right)},
\end{aligned}
\end{equation}
}
where $tp$ is temperature, a hyperparameter that controls the flatness of the distribution.  This equation measures the probability of target pixel-level instance $x_t^{(i)}$ being recognized as $k^\prime$-th source instance in our memory bank $\mathcal{B}$, acting as ``discriminate" in Figure \ref{overview}.
A similar calculation from the weakly augmented pixel-level sample $x_t^{\alpha(i)}$ can be defined as the instance-level pseudo-label:
{\small
\begin{equation}\label{weak_ins}
\begin{aligned}
& {y}_{ins}^{(i)} = [q_{1}^\alpha, \dots,q_{k^\prime}^\alpha, \dots], \\
& \textrm{where} \quad q_{k^\prime}^\alpha =\frac{\exp \left((e_t^\alpha)^{T} e_{k^\prime} / tp\right)}{\sum_{k=1}^{K} \exp \left((e_t^\alpha)^{T} e_{k} / tp\right)},
\end{aligned}
\end{equation}
}
Then, this naive instance-level pseudo-label will be used to adjust the semantic-level pseudo-label during the following pseudo-label regeneration process. Finally, an additional CE loss function is then introduced to minimize the difference between $\hat{y}_{ins}^{(i)}$ and ${y}_{ins}^{(i)}$:
{\small
\begin{eqnarray}\label{ins_pseudo}
\mathcal{L}_{ins}= -\sum_{i=1}^{H \times W} {y}_{ins}^{(i)} \log \left(\hat{y}_{ins}^{(i)}\right)
\end{eqnarray}
}
Finally, our overall UDA objective $\mathcal{L}_{overall}$ is  calculated as the weighted sum of each loss component as
$
\mathcal{L}_{{overall }}=\mathcal{L}_{src}+ \mathcal{L}_{tgt}+\lambda_{ins} \mathcal{L}_{ins}
$
, where $\lambda_{ins}$ is the parameter controlling the weight of instance loss.

\subsection{Pseudo-label Regeneration}
\label{interaction}
As we mentioned earlier, the naive semantic-level pseudo-labels are glutted with noises. To further improve the quality of the pseudo-labels, we creatively propose the pseudo-label regeneration to exhaustively utilize the labeled information and introduce a way to calibrate semantic predictions and instance predictions so that they could interact with and adjust each other. During the regeneration, our key objective is to align the semantic-level and instance-level predictions which possess different channel sizes. 

\iffalse
As illustrated in Figure \ref{pseudo-intercation}, if the predictions of both levels are highly similar, it means the prediction is mostly correct and should be enhanced (top panel). On the other hand, if the predictions are highly different, for example, the semantic-level recognizes the pixel as ``train'' while the instance-level regard it as the most similar to a ``car'' instance, it is likely that this ``disagreement'' indicates the occurrence of noisy labels (bottom panel). As a consequence, both predictions will be flattened and thus cause less harm to the model's performance.
\fi

\iffalse
\begin{figure}[t]
\centering  %图片全局居中
\includegraphics[width=0.7\linewidth]{fig/pseudo-interaction.pdf}
\caption{\normalsize Intuition behind pseudo-label regeneration. After we calibrate the predictions obtained from the two levels, we compare the highest possibility in each level. If the most possible predictions are similar within the two levels, the regenerated pseudo-label will become sharper in the most confident category. In contrast, if the predictions of  two levels disagree with each other, they will check and balance the interacted pseudo-labels, resulting in a much flatter distribution.} 
\label{pseudo-intercation}
\end{figure}
\fi

For an input weakly augmented target image $x^{\alpha}_t$ (bottom row of Figure \ref{overview}), we first extract its image-wise feature map $\mathbf{m}^{\alpha}_t$, then we obtain its semantic predictions using our pixel-level classifier $p^{\alpha}_t=g(\mathbf{m}^{\alpha}_t)$, $p^{\alpha}_t \in \mathbf{R}^{[H\times W] \times C}$.
%and instance-level similarity predictions of every pixel in $\mathbf{m}^{\alpha}_t$ via Eq. (\ref{weak_ins}) as $q^{\alpha} \in \mathbf{R}^{[1 \times 1] \times K}$. 
For a single pixel within $\mathbf{m}^{\alpha}_t$, we denote its semantic prediction as $z^{t} \in \mathbf{R}^{[1\times 1] \times C} $ and calculate its instance-level similarity predictions via Eq. (\ref{weak_ins}) as $q^{\alpha} \in \mathbf{R}^{[1 \times 1] \times K}$ .
Generally, $K$ is much larger than $C$ since at least one instance is needed for each semantic category. We then calibrate $z^t$ with $q^{\alpha}$ by  \textbf{scattering} $z^t$ into $K$ dimensional space, denoted as $z^{sc} \in \mathbf{R}^{[1 \times 1] \times K}$ 
% This is achieved by the \emph{gather} function with the expanded labels:
{\small
\begin{eqnarray}
\label{label_func}
z_{k}^{sc}=z_{j}^{t}, \text { if } {label}\left(q_{k}^{\alpha}\right)={label}\left(z_{j}^{t}\right),
\end{eqnarray}}
where $label(\cdot)$ is the function that returns the ground truth label. For example, $label(q_{k}^{\alpha})$ means the label for the $k^{th}$ element in the instance bank and ${label}\left(z_{j}^{t}\right)$ stands for the $j^{th}$ semantic category for the ``softmaxed'' prediction. 

% Figure \ref{scattering_pre} shows a intuitive example of how \textbf{scattering} works. 

\iffalse
\begin{figure}[t]
\centering  %图片全局居中
\includegraphics[width=0.8\linewidth]{fig/scattering_pre.pdf}
\caption{\normalsize The scattering process of semantic-level predictions.  Since the image-wise semantic prediction $z^t$ contains $H\times W$ pixels, we pick out prediction of a \emph{single pixel} to describe our scattering mechanism. The scattering process starts by matching the index (equals to the semantic class) of prediction with the ground truth category labels stored in our label bank $\mathcal{B}_l \in \mathbf{R}^{K \times 1}$, then maps the $C$ dimensional original semantic prediction values into the corresponding positions to form a vector with dimension $K$. Please note that the \emph{gathering} mechanism of instance-level predictions follows the same pattern but acts as a reversed process. } 
\label{scattering_pre}
\end{figure}
\fi

The regeneration of instance pseudo-labels is expressed as the \textbf{scaling} between new and old instance-level predictions:
\begin{eqnarray}
\hat{q}_{k}=\frac{q_{k}^{\alpha} z_{k}^{sc}}{\sum_{k=1}^{K} q_{k}^{\alpha} z_{k}^{{sc}}}.
\end{eqnarray}
We adopt the newly calibrated $\hat{q}$ to replace the old ${y}_{ins}$ in Eq.(\ref{ins_pseudo}). 

Furthermore, to interact the semantic-level with instance information, we \textbf{gather} $q$ into $C$ dimensional space by summing over instance predictions with shared labels
{\small
\begin{eqnarray}
q_{i}^{ga}=\sum_{j=0}^{K}
\mathbbm{1}\left({label}\left(z_{i}^{t}\right)={label}\left(q_{j}^{\alpha}\right)\right) q_{j}^{\alpha}
\end{eqnarray}
}
Now we may further adjust the semantic pseudo-label by \textbf{smoothing} $z^t$ with $q^{ga}$, written as:
{\small
\begin{eqnarray}\label{aggregate}
\hat{z}_{i}=\phi z_{i}^{t}+(1-\phi) q_{i}^{ga},
\end{eqnarray}
}
where $\phi$ is a hyper-parameter that balances the weight of semantic and instance information. Similarly, the adjusted $\hat{z}$ will replace the old semantic pseudo-label $\hat{y}_{t}$ in  Eq. (\ref{semantic-pseudo}). 
Note that the pseudo-regeneration of two levels starts simultaneously since we use copies of old ${y}_{ins}$ and $\hat{y}_{t}$ for regeneration.
By doing so, the dual-level information is subtly interacted with and follows the protocol that the distributions of two levels are always expected to agree with each other, as illustrated in Figure \ref{pseudo-intercation}.

\begin{figure}[t]
\centering  %图片全局居中
\includegraphics[width=0.7\linewidth]{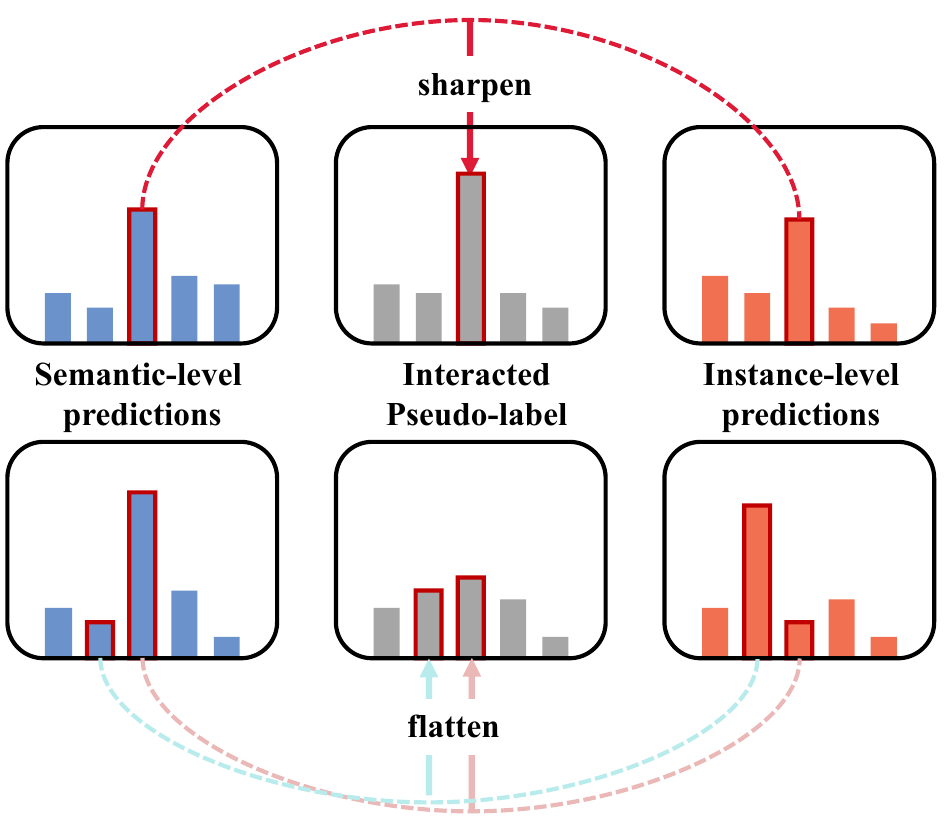}
\caption{\small Intuition behind pseudo-label regeneration. After we calibrate the predictions obtained from the two levels, we compare the highest possibility in each level. If the most possible predictions are similar within the two levels, the regenerated pseudo-label will become sharper in the most confident category. In contrast, if the predictions of  two levels disagree with each other, they will check and balance the interacted pseudo-labels, resulting in a much flatter distribution.} 
\label{pseudo-intercation}
\end{figure}

\subsection{Instance Bank and Updating Strategies}
\label{update-bank}
\iffalse
Different from contrastive learning for image classification \cite{instance-dis}, we view each pixel as one instance in the area of segmentation.
\fi
As mentioned above, with designed bank size $K$ and embedding size $D$, we construct a feature bank $\mathcal{B}_f \in \mathbf{R}^{K \times D}$ and a label bank $\mathcal{B}_l \in \mathbf{R}^{K \times 1}$ to keep instance-level information: the extracted instance embeddings and their corresponding ground truth labels. Different from domain adaptation for classification where features of all images can be stored, for the segmentation task, we strictly control the bank size and meticulously scheme its updating strategies.

% `` " are right format for quotation mark
Our key observation which motivates us to carefully design the updating strategy is that the long-tailed class distribution of source data (see \textcolor{orange}{supplementary material} for details) will result in a strong bias towards common classes (\eg, ``road", ``sky") instead of classes with very limited pixels (\eg, ``sign", ``light") and those only appear in a few samples (\eg, ``bike", ``rider"). In addition, most pixels of an entity are actually redundant and less determinant to the segmentation performance than those around object boundaries. To address these two issues and make the deposited instance feature embeddings more representative, we proposed two strategies, \ie, class-balanced sampling and boundary pixel selecting.

% ~\\ % empty line
\noindent
\textbf{Class-Balanced Sampling (CBS).} Before saving instance-level embeddings to $\mathcal{B}$, we set the same proportions of labeled place-holders for each class, \ie, $Holders = \frac{K}{C}$. %where $C$ is the number of classes. 
The feature bank is initiated offline with randomly selected instance embeddings but updated online through our selecting strategies.
% According to our observation, all blank spaces are quickly filled in with labeled instance embeddings during the very first training iterations(less than 10). 
We have also experimented with different bank sizes $K$ and different distributions of classes within, please see more details in our experiment and analysis (Sec. \ref{bankandstrategy}).

% ~\\ % empty line
\noindent
\textbf{Boundary Pixel Selecting (BPS).} %To filter and select boundary pixels, 
We first generate boundary masks $\mathcal{M}_s$ for each annotated source sample through Algorithm \ref{Boundary_mask},
then the boundary pixel maps $\mathcal{E}_s$ are easily calculated by $\mathcal{Y}_s * \mathcal{M}_s$.

\iffalse
\noindent
\textbf{Depositing Instances with EMA Smoothing.}
 At every update interval, $C$ embeddings are carefully selected from the input source image $X_{s}^{j}$ and mapped into the feature bank through their labels. 
                    % This part is not explicit!!!%
If $X_{s}^{j}$ does not contain a certain class of instances, we skip updating all embeddings belonging to this class for once.
Specifically, for the source domain samples, we first compute their feature maps and use our selecting techniques (which we will describe later) to filter $C$ instance embeddings $e^{slt} \in \mathbf{R}^{1 \times C}$ matching the $C$ semantic ground truth labels. Then these instances are deposited into our labeled instance bank via EMA smoothing after each update interval $u$:
\begin{eqnarray}\label{ema_s}
e_{u} \leftarrow \omega {e}_{u-1}+(1-\omega) e^{slt}
\end{eqnarray}
\fi

% \IncMargin{1em}
% \vspace{-5pt}

\begin{algorithm} {\small
\SetKwData{Left}{left}\SetKwData{This}{this}\SetKwData{Up}{up}
\SetKwFunction{Union}{Union}\SetKwFunction{FindCompress}{FindCompress} \SetKwInOut{Input}{Input}\SetKwInOut{Output}{Output}
	
	\Input{Ground truth label $\mathcal{Y}_s^{l}$ of size $H \times W$} 
	
	 %\BlankLine 
	 \Output{Boundary mask $\mathcal{M}_l$ of image $l$}
	 %\emph{special treatment of the first line}\; 
	 Initialization: All-zero matrix $\mathcal{M}_l$ of size $H \times W$, receptive field $R$ of size $3 \times 3$, threshold $\sigma \leftarrow 2$

	 \For{$i\leftarrow0$ \KwTo $H$}{ 
	 	%\emph{special treatment of the first element of line $i$}\; 
	 	\For{$j\leftarrow0$ \KwTo $W$}{
	 	Initiate $ClassCount\leftarrow0$
	 	
	 	$Current \leftarrow$ pixel $\mathcal{Y}[i, j]$
    ($Current$ is the position of current pixel)	
    
	 	$ClassCount$ $\leftarrow$ Count of different classes within $Current$ receptive field $R$
	 	
	 	\uIf{$ClassCount > \sigma$}{
	 	    $\mathcal{M}[i, j]\leftarrow1$
	 	}
 	 }
 	 } 
 	 	  \caption{  Boundary Mask Generation}
 	 	  \label{Boundary_mask} 
         }
 	 \end{algorithm} 

% \vspace{-15pt}
% \DecMargin{1em} 
For the $c$-th class,
we take the average of embeddings matching $\mathcal{E}_s^c$ as $emb_b^c$.
To balance our selection, we adopt K-means clustering \cite{k-means} on the non-edge instances and pick out the centroid embeddings $emb_\theta^c$ of the $c$-th cluster. Together, the averaged instance feature embeddings $emb^{avg} \in  \mathbf{R}^{1 \times C}$ between $emb_b$ and $emb_{\theta}$ are what we need to update the feature bank during each update interval $u$ with ema smoothing \cite{ema}:
{\small
\begin{eqnarray}\label{ema_equa}
emb_{u} \leftarrow \omega \cdot {emb}_{u-1}+(1-\omega) emb^{avg}
\end{eqnarray}
}
Here $\omega$ $\in$ [0, 1) is a momentum coefficient.
Unlike ProDA \cite{ProDA} which calculates target domain prototypes on the semantic pseudo-labels, our BPS obtains more representative source domain prototypes to perform instance-level discrimination.
% The embeddings of boundary pixels and centroid pixels merge together to form a more representative one.
Particularly, if the current image does not contain a certain class of instances, we skip updating all embeddings belonging to this class for once. 
%The overall updating strategy for bank $\mathcal{B}$ is illustrated in Algorithm \ref{updating}.
\iffalse
%\IncMargin{1em}
\begin{algorithm} %
\SetKwData{Left}{left}\SetKwData{This}{this}\SetKwData{Up}{up}
\SetKwFunction{Union}{Union}\SetKwFunction{FindCompress}{FindCompress} \SetKwInOut{Input}{Input}\SetKwInOut{Output}{Output}
	%At update interval $u$
	
	\Input{Source domain images $\mathcal{X}_s^{l}$ , Ground truth labels $\mathcal{Y}_s^{l}$ of size $H \times W$,
	
	} 
	
	 %\BlankLine 
	 
	 %\emph{special treatment of the first line}\; 
	 Initialization: All-zero matrix $\mathcal{M}_l$ of size $H \times W$, receptive field $R$ of size $3 \times 3$, threshold $\sigma \leftarrow 2$
	 
	 \For{$i\leftarrow0$ \KwTo $H$}{ 
	 	%\emph{special treatment of the first element of line $i$}\; 
	 	\For{$j\leftarrow0$ \KwTo $W$}{
	 	Initiate $ClassCount\leftarrow0$
	 	
	 	$Current \leftarrow$ pixel $\mathcal{Y}[i, j]$
	 	
	 	$ClassCount$ $\leftarrow$ Count of different classes within $R \odot Current$
	 	
	 	\uIf{$ClassCount > \sigma$}{
	 	    $\mathcal{M}[i, j]\leftarrow1$
	 	}
 	 }
 	 } 
 	 	  \caption{  Updating strategy for $\mathcal{B}$}
 	 	  \label{updating} 
 	 \Output{Boundary mask $\mathcal{M}_l$ of image $l$}
 	 \end{algorithm}
%\DecMargin{1em} 
\fi

\section{Experiments}

% comparison table1
\renewcommand\arraystretch{1}%保证每列高度是原先的1.5倍
\tabcolsep=5pt    % 调整列间距
\begin{table*}[t]\footnotesize\centering%星号表示双栏
\caption{ \small Comparison with state-of-the-art (SOTA) methods for UDA from GTA5 \cite{gta} $\rightarrow$ Cityscapes \cite{cs} adaptation. The results for DIDA are averaged over 3 random seeds. $^{\dagger}$ means we report the ProDA \cite{ProDA} and CPSL \cite{CPSL} results without further distillation for fair comparison. Noted that we use \colorbox{lightgray}{gray} to highlight 9 long-tailed classes, and symbols [$ ^{\star \circ \bullet}$] stand for analogous pairs. The best and second-best results are highlighted in \textbf{bold} and \underline{underline} font, respectively.}

% \begin{tabular}{m{ 1.65cm}<{\centering} c |c<{\centering} c c c c c c c c c c c c c c c c c c |m{0.5cm}<{\centering}}
  \begin{tabular}{m{1.80cm}<{\centering}m{1.2cm}<{\centering}|m{0.32cm}<{\centering}m{0.32cm}<{\centering}m{0.32cm}<{\centering}m{0.32cm}<{\centering}m{0.32cm}<{\centering}m{0.32cm}<{\centering}m{0.32cm}<{\centering}m{0.32cm}<{\centering}m{0.32cm}<{\centering}m{0.32cm}<{\centering}m{0.32cm}<{\centering}m{0.32cm}<{\centering}m{0.32cm}<{\centering}m{0.32cm}<{\centering}m{0.32cm}<{\centering}m{0.32cm}<{\centering}m{0.32cm}<{\centering}m{0.32cm}<{\centering}m{0.4cm}<{\centering}|m{0.5cm}<{\centering}}
 %设定每列的宽度以及对齐方式，并且可以做到自动换行
 
 % \rotatebox{90}{some rotated text} 
  \hline %\Xhline{1.0pt}%第一条粗线
        \textbf{Method} &  \textbf{Venue} & \rotatebox{90}{Road$ ^{\star}$} 
        & \rotatebox{90}{S.walk$ ^{\star}$} 
        & \rotatebox{90}{Build.} 
        & \rotatebox{90}{Wall} 
        & \rotatebox{90}{Fence} 
        & \rotatebox{90}{Pole} 
        & \rotatebox{90}{\cellcolor{lightgray}Light} 
        % \rotatebox{90}{\colorbox{yellow}{Light}}
        & \rotatebox{90}{\cellcolor{lightgray}Sign} 
        % \rotatebox{90}{\colorbox{yellow}{Sign}} 
        & \rotatebox{90}{Veget.} 
        & \rotatebox{90}{Terrain} 
        & \rotatebox{90}{Sky} 
        & \rotatebox{90}{\cellcolor{lightgray}Person} 
        % \rotatebox{90}{\colorbox{yellow}{Person}} 
        & \rotatebox{90}{\cellcolor{lightgray}Rider}  
        % \rotatebox{90}{\colorbox{yellow}{Rider}} 
        & \rotatebox{90}{Car} 
        & \rotatebox{90}{\cellcolor{lightgray}Truck$ ^{\circ}$} 
        % \rotatebox{90}{\colorbox{yellow}{Truck$ ^{\circ}$}} 
        & \rotatebox{90}{\cellcolor{lightgray}Bus$ ^{\circ}$} 
        % \rotatebox{90}{\colorbox{yellow}{Bus$ ^{\circ}$}} 
        & \rotatebox{90}{\cellcolor{lightgray}Train} 
        % \rotatebox{90}{\colorbox{yellow}{Train}} 
        & \rotatebox{90}{\cellcolor{lightgray}motor$ ^{\bullet}$}\; 
        % \rotatebox{90}{\colorbox{yellow}{motor$ ^{\diamond}$}\;} 
        &  \rotatebox{90}{\cellcolor{lightgray}Bike$ ^{\bullet}$}
        % \rotatebox{90}{\colorbox{yellow}{Bike$ ^{\diamond}$}} 
        & mIoU\\ 
  % \Xhline{0.8pt}%第二条粗线
  \hline
\end{tabular}

% second tabular
   \begin{tabular}{c}
   GTA5 $\rightarrow$ Cityscapes\\
  \end{tabular}

\iffalse
 \textbf{Method} & \tiny{Road} & \tiny{S.walk} & \tiny{Build.} & \tiny{Wall} & \tiny{Fence} & \tiny{Pole} &  \tiny{Light} & \tiny{Sign} & \tiny{Veget.} & \tiny{Terrain} & \tiny{Sky} & \tiny{Person} &  \tiny{Rider} & \tiny{Car} & \tiny{Truck} & \tiny{Bus} & \tiny{Train} &  \tiny{M.bike} &  \tiny{Bike} & mIoU\\ 
 \fi

 % \begin{tabular}{m{ 1.65cm}<{\centering} c |c<{\centering} c c c c c c c c c c c c c c c c c c |m{0.5cm}<{\centering}}
\begin{tabular}{m{1.80cm}<{\centering}m{1.2cm}<{\centering}|m{0.32cm}<{\centering}m{0.32cm}<{\centering}m{0.32cm}<{\centering}m{0.32cm}<{\centering}m{0.32cm}<{\centering}m{0.32cm}<{\centering}m{0.32cm}<{\centering}m{0.32cm}<{\centering}m{0.32cm}<{\centering}m{0.32cm}<{\centering}m{0.32cm}<{\centering}m{0.32cm}<{\centering}m{0.32cm}<{\centering}m{0.32cm}<{\centering}m{0.32cm}<{\centering}m{0.32cm}<{\centering}m{0.32cm}<{\centering}m{0.32cm}<{\centering}m{0.4cm}<{\centering}|m{0.5cm}<{\centering}}

 % \Xhline{0.8pt}%第二条粗线
 \hline
        DACS \cite{DACS} & WACV'21 & 89.9 & 39.7 & 87.9& 30.7& 39.5& 38.5& 46.4& 52.8& 88.0& 44.0& 88.8& 67.2& 35.8& 84.5& 45.7& 50.2& 0.0& 27.3 & 34.0 & 52.1\\
        BiMAL \cite{COM-BIMAL} & ICCV'21 & 91.2 & 39.6 & 82.7 & 29.4 & 25.2& 29.6 & 34.3& 25.5 & 85.4& 44.0& 80.8& 59.7& 30.4& 86.6& 38.5& 47.6& 1.2& 34.0 & 36.8 & 47.3\\
        UncerDA \cite{COM-UNSERDA} & ICCV'21 & 90.5 & 38.7 & 86.5 & 41.1 & 32.9& 40.5 & 48.2& 42.1 & 86.5& 36.8& 84.2& 64.5& 38.1& 87.2& 34.8& 50.4& 0.2& 41.8 & 54.6 & 52.6\\
        BAPA-Net \cite{BAPA-Net} & ICCV'21 & 94.4 &  61.0 &  88.0 & 26.8 & 39.9 & 38.3 & 46.1 & 55.3 &  87.8 & 46.1 &  89.4 &  68.8 &  40.0 & 90.2 &  60.4 &  59.0 &  0.00 & 45.1 &  54.2 & 57.4\\
        DPL-Dual \cite{COM-DPL-DUAL} & ICCV'21 & 92.8 & 54.4 & 86.2 & 41.6 & 32.7& 36.4 & 49.0& 34.0 & 85.8& 41.3& 86.0& 63.2& 34.2& 87.2& 39.3& 44.5& 18.7& 42.6 & 43.1 & 53.3\\
        UPLR \cite{UPLR} & ICCV'21 & 90.5 &  38.7 & 86.5 & 41.1 & 32.9 & 40.5 & 48.2 & 42.1 & 86.5 & 36.8 & 84.2 & 64.5 & 38.1 & 87.2 & 34.8 & 50.4 & 0.2 & 41.8 & 54.6 & 52.6 \\
        ProDA$ ^{\dag}$ \cite{ProDA} & CVPR'21 &  91.5 &  52.4 &  82.9 & 42.0 &  35.7 & 40.0 & 44.4 & 43.3 & 87.0 &  43.8 &  79.5 & 66.5 & 31.4 &  86.7 &  41.1 &  52.5 &  1.0 &  45.4 & 53.8  & 53.7 \\
        SAC \cite{SAC}  & CVPR'21 & 90.4 & 53.9 & 86.6 & 42.4 & 27.3 & 45.1 & 48.5 & 42.7 & 87.4 & 40.1 & 86.1 & 67.5 & 29.7 & 88.5 & 49.1 & 54.6 & 9.8 & 26.6 & 45.3 & 53.8 \\
        % CorDA \cite{corda} & 94.7& 63.1& 87.6& 30.7 & 40.6 & 40.2 &   47.8&  51.6 &  87.6 & 47.0 & 89.7 & 66.7 & 35.9 & 90.2 & 48.9 & 57.5 &  0.0 &   39.8&  56.0 & 56.6 \\
        %ProDA \cite{ProDA}  &  87.8 &  56.0 & 79.7 & 46.3 & 44.8 & 45.6 & 53.5 & 53.5 & 88.6 & 45.2 &   82.1& 70.7 & 39.2 & 88.8 & 45.5 & 59.4 & 1.0 &  48.9 & 56.4  & 57.5 \\
        % CPSL \cite{CPSL}  &  92.3 &  59.9 & 84.9 & 45.7 & 29.7& 52.8 & \textbf{61.5} &  59.5 & 87.9 & 41.5 & 85.0 &  73.0 & 35.5 & 90.4 & 48.7 & 73.9 & 26.3 & 53.8 & 53.9 & 60.8 \\
% model -- 2022+----
        % RPLR \cite{RPLR}  & TPAMI'22 & 92.3 & 52.3 & 84.8 & 34.7 & 29.7& 32.6 &36.7 & 32.7 & 83.2& 42.5& 81.5& 60.6& 33.3& 85.0& 44.2& 48.0& 3.8& 35.7 & 37.3 & 50.1\\
        CPSL$ ^{\dag}$ \cite{CPSL} & CVPR'22 &  91.7 &  52.9 & 83.6 & 43.0 & 32.3 & 43.7 & 51.3 &  42.8 &  85.4 &  37.6 & 81.1 &  69.5 & 30.0 & 88.1 & 44.1 & 59.9 & 24.9 &  47.2 & 48.4 & 55.7 \\
        CaCo \cite{caco} & CVPR'22 & 93.8 & 64.1 &  85.7 &  43.7 &  42.2 &  46.1 &  50.1  & 54.0 &  88.7 & 47.0 &  86.5 &  68.1 &  2.9 & 88.0 & 43.4 & 60.1 & 31.5 &  46.1 &  60.9 & 58.0 \\
        % ProCA \cite{ProCA} & 91.9 & 48.4 &  87.3 &  41.5 & 31.8 &  41.9 & 47.9 & 36.7 &  86.5 & 42.3 &  84.7 &  68.4 & 43.1 & 88.1 & 39.6 & 48.8 & 40.6 & 43.6 & 56.9 & 56.3\\
        % FAFS \cite{FAFS} & 93.4 & 60.7 & 88.0 & 43.5 & 32.1 & 40.3 & 54.3 & 53.0 & 88.2 & 44.5 & 90.0 & 69.5 & 35.8 & 88.7 & 34.1 & 53.9 & 41.3 & 51.7 & 54.7 & 58.8 \\
        % A &  &  &  &  &  &  &  &  &  &  &  &  &  &  &  &  &  &  &  &  \\
        % A &  &  &  &  &  &  &  &  &  &  &  &  &  &  &  &  &  &  &  &  \\
        % G2L \cite{G2L} & 95.8 &  68.8 & 88.0 & 46.5 & 37.5 & 50.3 & 58.4 & 58.1 & 89.5 & 51.5 & 83.1 & 69.0 & 33.6 & 89.6 & 41.3 & 59.4 & 36.9 & 46.9 & 29.9 & 59.7 \\
        DAFormer \cite{DAFormer} & CVPR'22 & \underline{95.7}  & \underline{70.2} & \underline{89.4}  & \underline{53.5} & \underline{48.1} & \underline{49.6} &  \underline{55.8} &  \underline{59.4} & \underline{89.9} &  \underline{47.9}  & \underline{92.5} & \underline{72.2} &  \underline{44.7} &  \underline{92.3} & \underline{74.5} & \underline{78.2} & \underline{65.1} &  \underline{55.9} &  \underline{61.8} & \underline{68.3} 
        % \\
        % ARAS \cite{com-ARAS}  & TCSVT'23 & 91.9 & 45.2 & 81.8 & 21.9 & 25.6& 35.5 & 41.5& 33.4 & 85.1& 34.8& 73.8& 62.5& 31.6& 85.9& 33.8& 42.5& 7.3& 33.8 & 42.8 & 47.9
        % CAMix \cite{classmix} & TCSVT'22 & \underline{96.0} & \underline{73.1} & \underline{89.5} & \underline{53.9} & \underline{50.8} & \underline{51.7} & \underline{58.7} & \underline{64.9} & \underline{90.0} & \underline{51.2} & \underline{92.2} & \underline{71.8} & \underline{44.0} & \underline{92.8} & \underline{78.7} & \underline{82.3} & \underline{70.9} & \underline{54.1} & \underline{64.3} & \underline{70.0}
        \\
        \textbf{DIDA (Ours)} & – & \textbf{97.3}  & \textbf{78.0} & \textbf{89.8} & \textbf{55.9} & \textbf{52.6} & \textbf{53.3} & \textbf{57.9} & \textbf{66.1} & \textbf{90.0} & \textbf{50.0} & \textbf{93.1} & \textbf{73.2} & \textbf{44.8} & \textbf{93.4} & \textbf{80.9} & \textbf{84.7} & \textbf{73.8} & \textbf{58.6} & \textbf{63.5}  & \textbf{71.0} \\
\iffalse
        DACS \cite{DACS}  & Road & S.walk & Build. & Wall & Fence & Pole &  Light & Sign & Veget. & Terrain & Sky & Person &  Rider & Car & Truck & Bus & Train &  M.bike &  Bike & mIoU \\ 
\fi
  % \Xhline{0.8pt}%第三条粗线
  \hline
  \end{tabular}

    \label{comparison}
\end{table*}

% --------------------------------------------------
% comparison table2
% --------------------------------------------------
\renewcommand\arraystretch{1}%保证每列高度是原先的1.5倍
\tabcolsep=5pt    % 调整列间距
\begin{table*}[t]\footnotesize\centering%星号表示双栏
\caption{ \small Comparison with state-of-the-art (SOTA) methods from SYNTHIA \cite{synthia} $\rightarrow$ Cityscapes \cite{cs} adaptation.  mIoU* denotes the mean IoU over 13 classes excluding ``wall", ``fence", and ``pole".}

% \begin{tabular}{m{ 1.65cm}<{\centering} c |c<{\centering} c c c c c c c c c c c c c c c c c c |m{0.5cm}<{\centering}}
  \begin{tabular}{m{1.80cm}<{\centering}m{1.2cm}<{\centering}|m{0.36cm}<{\centering}m{0.36cm}<{\centering}m{0.36cm}<{\centering}m{0.36cm}<{\centering}m{0.36cm}<{\centering}m{0.36cm}<{\centering}m{0.36cm}<{\centering}m{0.36cm}<{\centering}m{0.36cm}<{\centering}m{0.36cm}<{\centering}m{0.36cm}<{\centering}m{0.36cm}<{\centering}m{0.36cm}<{\centering}m{0.36cm}<{\centering}m{0.36cm}<{\centering}m{0.4cm}<{\centering}|m{0.55cm}<{\centering}m{0.60cm}<{\centering}}
 %设定每列的宽度以及对齐方式，并且可以做到自动换行
 
 % \rotatebox{90}{some rotated text} 
  \hline %\Xhline{1.0pt}%第一条粗线
        \textbf{Method} 
        &  \textbf{Venue} 
        & \rotatebox{90}{Road$ ^{\star}$} 
        & \rotatebox{90}{S.walk$ ^{\star}$} 
        & \rotatebox{90}{Build.} 
        & \rotatebox{90}{Wall} 
        & \rotatebox{90}{Fence} 
        & \rotatebox{90}{Pole} 
        & \rotatebox{90}{\cellcolor{lightgray}Light} 
        % \rotatebox{90}{\colorbox{yellow}{Light}}
        & \rotatebox{90}{\cellcolor{lightgray}Sign} 
        % \rotatebox{90}{\colorbox{yellow}{Sign}} 
        & \rotatebox{90}{Veget.} 
        % & \rotatebox{90}{Terrain} ---no such class
        & \rotatebox{90}{Sky} 
        & \rotatebox{90}{\cellcolor{lightgray}Person} 
        % \rotatebox{90}{\colorbox{yellow}{Person}} 
        & \rotatebox{90}{\cellcolor{lightgray}Rider}  
        % \rotatebox{90}{\colorbox{yellow}{Rider}} 
        & \rotatebox{90}{Car} 
        % & \rotatebox{90}{\cellcolor{lightgray}Truck$ ^{\circ}$} ---no such class
        % \rotatebox{90}{\colorbox{yellow}{Truck$ ^{\circ}$}} 
        & \rotatebox{90}{\cellcolor{lightgray}Bus} 
        % \rotatebox{90}{\colorbox{yellow}{Bus$ ^{\circ}$}} 
        % & \rotatebox{90}{\cellcolor{lightgray}Train} ---no such class 
        % \rotatebox{90}{\colorbox{yellow}{Train}} 
        & \rotatebox{90}{\cellcolor{lightgray}motor$ ^{\bullet}$}\; 
        % \rotatebox{90}{\colorbox{yellow}{motor$ ^{\diamond}$}\;} 
        &  \rotatebox{90}{\cellcolor{lightgray}Bike$ ^{\bullet}$}
        % \rotatebox{90}{\colorbox{yellow}{Bike$ ^{\diamond}$}} 
        & mIoU & mIoU*\\ 
  % \Xhline{0.8pt}%第二条粗线
  \hline
\end{tabular}

   \begin{tabular}{c}
   SYNTHIA $\rightarrow$ Cityscapes\\
  \end{tabular}

\begin{tabular}{m{1.80cm}<{\centering}m{1.2cm}<{\centering}|m{0.36cm}<{\centering}m{0.36cm}<{\centering}m{0.36cm}<{\centering}m{0.36cm}<{\centering}m{0.36cm}<{\centering}m{0.36cm}<{\centering}m{0.36cm}<{\centering}m{0.36cm}<{\centering}m{0.36cm}<{\centering}m{0.36cm}<{\centering}m{0.36cm}<{\centering}m{0.36cm}<{\centering}m{0.36cm}<{\centering}m{0.36cm}<{\centering}m{0.36cm}<{\centering}m{0.4cm}<{\centering}|m{0.55cm}<{\centering}m{0.60cm}<{\centering}}
 
 \hline
        DACS \cite{DACS} & WACV'21 &  80.6 &  25.1 & 81.9 & 21.5 & 2.9 & 37.2 &  22.7 &  24.0 & 83.7   & \underline{90.8} & 67.6 &  38.3 &  82.9  & 38.9  &  28.5&  47.6  & 48.3 & 54.8 \\
        BiMAL \cite{COM-BIMAL} & ICCV'21 & \textbf{92.8} & \underline{51.5} & 81.5 & 10.2 & 1.0 & 30.4 & 17.6 & 15.9 & 82.4  & 84.6 & 55.9 & 22.3 & 85.7  & 44.5  & 24.6 & 38.8 & 46.2 & 53.7\\
        UncerDA \cite{COM-UNSERDA} & ICCV'21 & 79.4 & 34.6 & 83.5 & 19.3 & 2.8 & 35.3 & 32.1 & 26.9 & 78.8  & 79.6 & 66.6 & 30.3 & 86.1  &36.6   & 19.5 & 56.9 & 48.0 & 54.6\\
        BAPA-Net \cite{BAPA-Net} & ICCV'21 & 91.7 &  53.8 & 83.9 & 22.4 &  0.8 &  34.9 & 30.5 & 42.8 & 86.6  & 88.2 & 66.0 & 34.1 & 86.6  & 51.3 & 29.4 & 50.5 & 53.3 & 61.2\\
        DPL-Dual \cite{COM-DPL-DUAL} & ICCV'21 & 83.5 & 38.2 & 80.4 & 1.3 & 1.1 & 29.1 & 20.2 & 32.7 & 81.8  & 83.6 & 55.9 & 20.3 & 79.4  & 26.6  & 7.4 & 46.2 & 43.0 & 50.5\\
        UPLR \cite{UPLR} & ICCV'21 & 79.4 & 34.6 & 83.5 & 19.3 & 2.8 & 35.3 & 32.1 & 26.9 & 78.8  & 79.6 & 66.6 & 30.3 & 86.1 & 36.6 & 19.5 & 56.9 & 48.0 & 54.6\\
        ProDA$ ^{\dag}$ \cite{ProDA}  & CVPR'21 & 87.3 & 44.0 &  83.2 & 26.9 & 0.7 &   42.0 &  45.8 & 34.2 & 86.7  & 81.3 & 68.4 & 22.1 & 87.7  &  50.0  &  31.4 &  38.6 & 51.9 & 62.0\\
        % CorDA \cite{corda}  &  \textbf{93.3} & \textbf{61.6} & 85.3 & 19.6 &  5.1 & 37.8 &  36.6 &  42.8 & 84.9 &  – & 90.4 & 69.7 &  41.8 & 85.6 &  – &  38.4 &  – &   32.6 &  53.9 & 55.0 \\
        SAC \cite{SAC} & CVPR'21 & 89.3 & 47.2 & 85.5 & 26.5 & 1.3 & 43.0 & 45.5 & 32.0 & \underline{87.1}  & 89.3 & 63.6 & 25.4 & 86.9  & 35.6  & 30.4 & 53.0 & 52.6 &  59.3\\
% model -- 2022+----
        % ARAS \cite{com-ARAS}  & TCSVT'23 & 85.6 & 39.2 & 79.9 & 15.5 & 0.3& 32.2 & 19.3& 23.9 &79.1 &81.7 & 61.1& 19.3& 82.9& 25.7& 10.6& 51.9& 44.3 &  &  & \\
        % RPLR \cite{RPLR}  & TPAMI'22 & 81.5 & 36.7 & 78.6 & 1.3 & 0.9 & 32.2 & 20.7 & 23.6 & 79.1  & 83.4 & 57.6 & 30.4 & 78.5  & 38.3  & 24.7 & 48.4 & 44.7\\
        % ProCA \cite{ProCA}& ECCV'22 & 90.5 & 52.1 & 84.6 & 29.2 &  3.3 & 40.3 & 37.4 & 27.3 & 86.4  & 85.9 & 69.8 & 28.7 & 88.7  & 53.7  & 14.8 & 54.8 & 53.0\\
        CPSL$ ^{\dag}$ \cite{CPSL}  & CVPR'22 & 87.3 &  44.4 &  83.8 &  25.0 &  0.4 &  42.9 &  47.5 & 32.4 & 86.5  & 83.3 &  69.6 & 29.1 & \textbf{89.4}  &  52.1  & 42.6 & 54.1 & 54.4 & 61.7 \\
        CaCo \cite{caco} & CVPR'22 & 87.4 & 48.9 & 79.6 &  8.8 & 0.2 &  30.1 &  17.4 & 28.3 & 79.9  & 81.2 &  56.3 &  24.2 & 78.6  & 39.2 &  28.1 & 48.3 &  46.0 & 53.6\\
        % FAFS \cite{FAFS} & 83.1 & 43.4 & 83.7 & 29.3 & 0.9 & 44.2 & 46.1 & 38.1 & 86.6 & – & \underline{91.1} & 67.4 & 28.1 & \underline{88.6} & – & 45.9 & – & 37.1 & 58.7 & 54.5 \\
        % A &  &  &  &  &  &  &  &  &  & – &  &  &  &  & – &  & – &  &  &  \\
        % A &  &  &  &  &  &  &  &  &  & – &  &  &  &  & – &  & – &  &  &  \\
        % G2L \cite{G2L} & 87.8 & 45.1 & 85.1 & 25.2 & 1.3 & 46.9 & 53.3 & 46.4 & 88.1 & – & 86.0 & 72.0 & 40.6 & 91.4 & – & 62.9 & – & 35.6 & 41.1 & 56.8 \\
        % ProDA \cite{ProDA}  &  \textbf{87.8} & 45.7 & 84.6 &  37.1 & 0.6 &  44.0 & 54.6 & 37.0 & \textbf{88.1} & – & 84.4 & 74.2 & 24.3 & \textbf{88.2} & – &  51.1 & – & 40.5 &  45.6 & 55.5 \\
        % CPSL \cite{CPSL}  & 87.2  &  43.9 & 85.5 & 33.6 &  0.3 &  47.7 &  \textbf{57.4} & 37.2 & 87.8 & – & 88.5 &  \textbf{79.0} & 32.0 & \textbf{90.6} & – & 49.4 & – & 50.8 & 59.8 & 57.9   \\
        DAFormer \cite{DAFormer}  & CVPR'22 & 84.5 & 40.7 &  \underline{88.4} & \underline{41.5} &  \underline{6.5} &  \underline{50.0} & \underline{55.0} & \underline{54.6} & 86.0  & 89.8 & \underline{73.2} & \underline{48.2} & 87.2   & \underline{53.2}  & \underline{53.9} &  \underline{61.7} & \underline{60.9} & \underline{67.4}\\
        % ARAS \cite{com-ARAS}  & TCSVT'23 & 85.6 & 39.2  & 79.9 & 15.5 & 0.3 & 32.2 & 19.3 & 23.9 & 79.1  & 81.7 & 61.1 & 19.3 & 82.9  & 25.7 & 10.6 & 51.9 & 44.3 \\
        \textbf{DIDA (Ours)} & – &  \underline{90.6} & \textbf{53.7} & \textbf{88.5} & \textbf{45.7} & \textbf{8.5} & \textbf{50.5} & \textbf{56.8} & \textbf{56.1} & \textbf{87.8}  & \textbf{91.5} & \textbf{74.6} & \textbf{49.6} & \underline{88.1}   & \textbf{62.7}  & \textbf{56.2} & \textbf{64.3}  & \textbf{63.3} & \textbf{70.1} \\
  % \Xhline{1.0pt}%第三条粗线
  \hline
  \end{tabular}
  
    \label{comparison1}
\end{table*}

\subsection{Implementation Details}
\textbf{Training.}
We use the mmsegmentation framework\footnote{https://github.com/open-mmlab/mmsegmentation} with backbone \cite{segformer}, which is pre-trained on ImageNet. Strictly following DAFormer \cite{DAFormer}, we utilize Rare Class Sampling, Thing-Class ImageNet Feature Distance, and Learning Rate Warmup with the same settings for its hyper-parameters.
In accordance with \cite{DACS}, we apply color jitter, Gaussian blur, and ClassMix as data augmentations. Our model is trained on a batch of
two 512×512 random crops for 40k iterations.

\noindent
\textbf{Datasets.} We conduct our experiments on the two widely-used UDA benchmarks, \ie, GTA5 \cite{gta} $\rightarrow$ Cityscapes \cite{cs} and SYNTHIA \cite{synthia} $\rightarrow$ Cityscapes \cite{cs}. We report the results of 19 shared categories for GTA5 and 16 common categories for SYNTHIA with Cityscapes.
Noted that we use an identical set of hyper-parameters for both datasets ($K=300$, $\lambda_{ins} = 1$, $tp=0.1$, $\tau=0.968$, $\phi=0.9$, $\omega=0.999$, $u=50$).

\subsection{Comparison with the SOTA methods}
% Comparisons with state-of-the-art methods}
% We exhaustively compare our proposed method with 5 recently leading approaches.

% As shown in Table \ref{comparison}, our proposed method outperforms the previously leading approaches by a large margin. 

Table \ref{comparison} and \ref{comparison1} demonstrate the effectiveness of our proposed method in UDA. It outperforms all other baselines developed in the recent three years and presented in top-tier conferences and journals.

\noindent
\textbf{Results of GTA5 $\rightarrow$ Cityscapes.}
For GTA5$\rightarrow$Cityscapes adaptation (shown in Table \ref{comparison}), DIDA achieves the best IoU score in all 19 categories, and it attains the state-of-the-art mIoU score of 71.0, surpassing the second-best method \cite{DAFormer} by a large margin of \textbf{2.7}. This can be ascribed to the further exploration of instance-level information and pseudo-label regeneration.
Owing to the boundary pixel selecting strategy, our model learns to accurately distinguish between adjacent and analogous instances (\eg, ``sidewalk" and ``road", ``truck" and ``bus"), thus promoting the segmentation performance significantly (``sidewalk" by \textbf{7.8} percent and ``truck" by \textbf{6.4}).
It is also worth mentioning that DIDA shows prominent advantages in handling the long-tailed categories, such as improving the IoU score of ``train" by \textbf{8.7}, and ``sign" by \textbf{6.7}.

\noindent
\textbf{Results of SYNTHIA $\rightarrow$ Cityscapes.}
As displayed in Table \ref{comparison1}, we still observe significant improvements over competing methods on the more challenging SYNTHIA $\rightarrow$ Cityscapes benchmark. Specifically, DIDA tops over 14 among all 16 categories while  achieving the best mIoU performance of 63.3, outperforming the second-best method DAFormer \cite{DAFormer} by \textbf{2.4}. It is worth noticing that DIDA's capability to distinguish between analogous pairs works particularly well on the SYNTHIA$\rightarrow$Cityscapes benchmark. Similarly, on confusing classes like ``road" and ``sidewalk", DIDA improves the baseline \cite{DAFormer} model's performance by a significant margin (road by \textbf{6.1}, sidewalk by \textbf{13.0}). %\yao{Mention that DAFormer baseline model lacks refinement on road and sidewalk, our model improve its performance by a significant margin, road by 6.1\%, sidewalk by 13\%. }
These results also reveal the effectiveness of our DIDA among some of the hardest categories, \eg, ``fence",  ``sign", and ``motor".

\begin{figure}[t]
\centering  %图片全局居中
\subfigure[Bank Size]{
\label{k}
\includegraphics[width=3.92cm,height = 2.8cm]{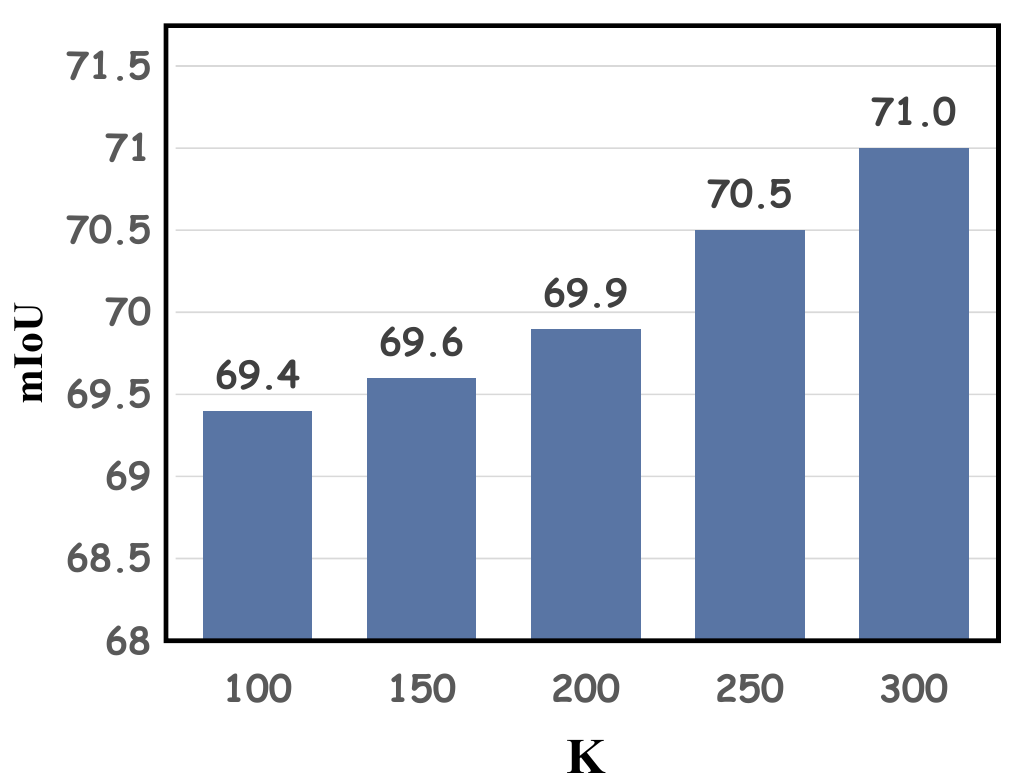}}\subfigure[Update Interval]{
\label{u}
\includegraphics[width=3.92cm,height = 2.8cm]{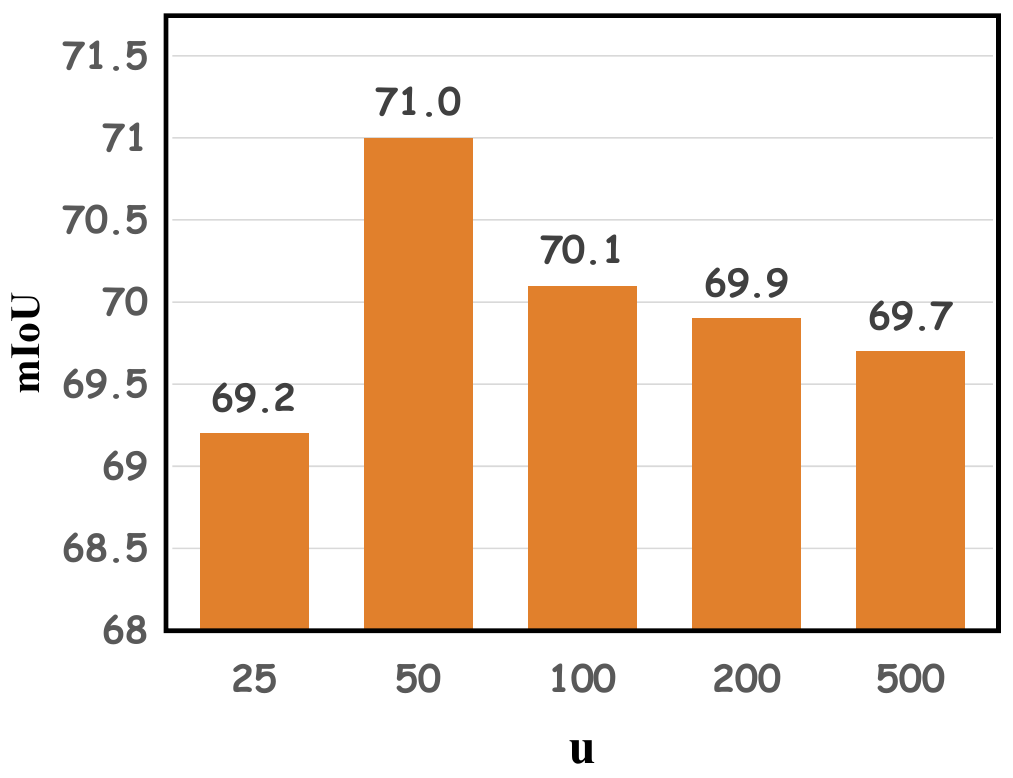}}
% \vspace{-15pt}
\caption{ \normalsize Results of varying $K$ and $u$.}
\label{1}
\end{figure}

\iffalse
\begin{figure}
\centering  %图片全局居中
\includegraphics[width=0.8\linewidth]{fig/sup-lt2.pdf}
\caption{\normalsize Category distribution of GTA5 \cite{gta}}
\label{longtail}
\end{figure}
\fi

\subsection{Analysis of Instance Feature Selection}
\label{bankandstrategy}
The success of our DIDA largely lies in the introduction of instance-level discrimination, which can be ascribed to our meticulously designed dynamic memory bank mechanism and the proper selection of more informative source domain instances. Herein, we analyze how the bank size and different updating schemes can affect the effectiveness of DIDA.

\noindent
\textbf{Choice of Bank Size.}
As a general rule of thumb, the more instance features are stored, the model's performance improves. We present results of varying different bank sizes $K$ in Figure \ref{k}, which validates this claim. Note that when $K$ surpasses 300, the consumption of storage increases to more than 23 GB, which pushes the GPU memory to its limit (24GB for a single RTX 3090 GPU). 
By limiting the bank size, our model is both friendly in memory occupation and efficient in adaptation. In the evaluation and the following ablation studies, our bank size $K$ is set to $300$.

\noindent
\textbf{Updating Strategies.} 
This part includes experiments of update interval $u$, sampling strategy, and selecting strategy. Specifically, we sweep over [25, 50, 100, 200, 500] for update interval $u$. It's obvious in Figure \ref{u} that $u=50$ achieves the top result, while $u=25$ results in the worst. This is consistent with what MoCo \cite{moco} points out: a quickly changing feature bank leads to a dramatic reduction in performance. 
For updating features in the bank, we investigate no update (NU), random sampling (RS), inverted long-tail sampling (ILS), and class-balanced sampling (CBS). We also use AVG, KM, and BPS to represent average, K-means clustering, and boundary pixel selecting strategies. Please be aware that RS results in the long-tail distribution of classes and ILS leads to  a “polar” condition of long-tail distribution (the original “head” becomes the new “tail”). %where the original “head” classes become the new “tail” classes. % , namely, the “Inverted” long-tail sampling.
Results are shown in Table \ref{update}.

\begin{figure*}
\centering
\includegraphics[width=1\linewidth]{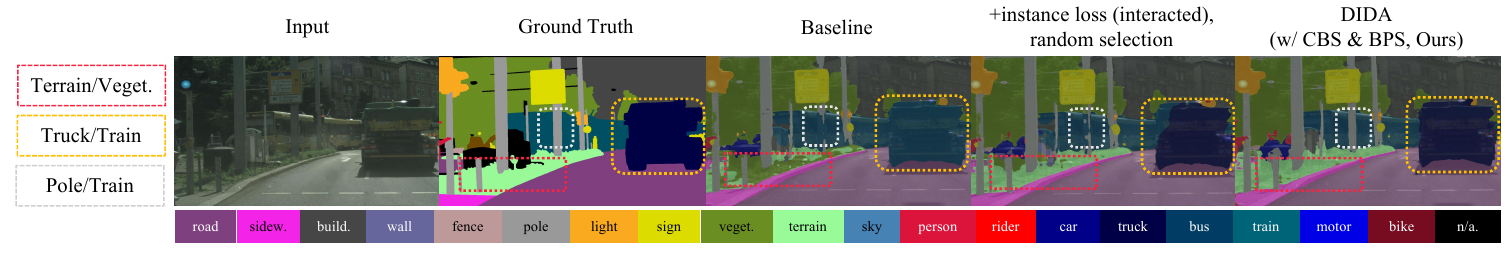}
\caption{\normalsize Visualized ablation of Sampling strategies.}
\label{fig:comparison}
\end{figure*}

% ablation 2
\renewcommand\arraystretch{1}%保证每列高度是原先的1.5倍
\begin{table}[h]\centering%星号表示双栏
% \vspace{-15pt}
\caption{\normalsize Ablation on sampling and selecting strategies}
% \vspace{5pt}
\scalebox{0.9}{
 \begin{tabular}{c|c|c|c} %设定每列的宽度以及对齐方式，并且可以做到自动换行
 
 % \rotatebox{90}{some rotated text} 
% \Xhline{0.8pt}%第一条粗线
        \textbf{Sampling}  & mIoU &\textbf{Selecting} & mIoU\\ 
% \Xhline{1.0pt}%第一条粗线
% \midrule[1.0pt]
\hline
       NU & 68.6 & RS & 69.5 \\ 
       RS & 69.5 & AVG & 69.7 \\ 
       ILS & 69.0 & KM & 69.6 \\ 
        CBS & \textbf{71.0} & BPS & \textbf{71.0} \\ 
% \Xhline{0.8pt}
  % \Xhline{0.8pt}%第二条粗线
\end{tabular}
}
    \label{update}
\end{table}

\subsection{Ablation Studies}
\label{ablation study}
All ablation studies are conducted on the GTA5 $\rightarrow $ Cityscapes dataset. For more ablations, please refer to our \textcolor{orange}{supplementary material}.

\noindent
\textbf{Smooth Parameters.}
Table \ref{smooth} reveals the effect of \emph{gathering} smooth parameters $\phi$ in Eq.(\ref{aggregate}). 
It indicates that the best proportion of semantic information locates at 0.9. We consider the extreme conditions as well: $\phi=1$ equals taking the original semantic pseudo-label for Eq.(\ref{semantic-pseudo}), while when $\phi=0$, the $\mathcal{L}_{tgt}$ oscillates and fails to converge.

% \vspace{-20pt}
% ablation 3 in supp
\renewcommand\arraystretch{1}%保证每列高度是原先的1.5倍
\begin{table}[h]\centering%星号表示双栏
\caption{\normalsize Results of different smooth parameters $\phi$}
% \vspace{3pt}
\scalebox{0.9}{
 \begin{tabular}{c|c|c|c|c|c} %设定每列的宽度以及对齐方式，并且可以做到自动换行
 
 % \rotatebox{90}{some rotated text} 

       \textbf{$\phi$}  & 0 & 0.8 & 0.9 & 0.95 & 1.0\\ 
% \Xhline{1.2pt}%第一条粗线
\hline
       mIoU  & \emph{fail} & 70.2 & \textbf{71.0} & 70.3 & 69.0\\

  % \Xhline{0.8pt}%第二条粗线
\end{tabular}
}
    \label{smooth}
\end{table}

\noindent
\textbf{Pseudo-label regeneration Strategies.}
We also tried different combinations to find out the most suitable regeneration strategy to regenerate instance-level pseudo-label $\hat{q}$ and semantic-level pseudo-label $\hat{z}$. As observed from Table \ref{interaction_stg}, applying \textbf{smoothing} to $\hat{z}$ and \textbf{scaling} to $\hat{q}$ obtains the top result. While there is very little difference from applying smoothing to the pseudo-labels from both levels (also achieves 70.9 mIoU), another smoothing parameter is then introduced in the process. Therefore, we prefer the scaling for $\hat{q}$ to keep a simpler framework

\renewcommand\arraystretch{1}%保证每列高度是原先的1.5倍
\begin{table}[h]\centering%星号表示双栏
\caption{\normalsize Results of different interacting strategies $\phi$}
% \vspace{3pt}
\scalebox{1.0}{
\begin{tabular}{|p{0.2\linewidth}|m{0.18\linewidth}<{\centering}|m{0.15\linewidth}<{\centering}|}
  \hline
  \diagbox[width=5.8em, height=2em]{$\hat{z}$}{$\hat{q}$} & Smoothing & Scaling \\
  \hline
  Smoothing & 70.9 & \textbf{71.0} \\
  \hline
  Scaling & 70.5 & 70.3 \\
  \hline
\end{tabular}
}
    \label{interaction_stg}
\end{table}

% \yao{design a cross table of the interaction strategies of dual-level predictions. scaling for q and smoothing for z}

%---------------------additional ablations--------------------
\noindent
\textbf{EMA Momentum.}
Table \ref{ema} shows mIoU performance with different ema momentum $\omega$ in Eq.(\ref{ema_equa}).
%\begin{eqnarray}
%e_{u} \leftarrow \omega {e}_{u-1}+(1-\omega) e^{avg}
%\end{eqnarray}
It performs reasonably from 0.9 to 0.999, demonstrating that a relatively slow-updating (\ie, larger momentum) instance bank is effective. Under the extreme condition $\omega=0$, the instance bank is purely updated with newly selected instance features, and the performance reduces dramatically. When $\omega=1$, it equals to ``No Update" as conducted in the analysis of ``Sampling Strategies".

% ablation ema
\renewcommand\arraystretch{1}%保证每列高度是原先的1.5倍
\begin{table}[h]\centering%星号表示双栏
\caption{\normalsize Results of different ema momentum $\omega$}
\scalebox{0.9}{
 \begin{tabular}{c|c|c|c|c|c} %设定每列的宽度以及对齐方式，并且可以做到自动换行
 
 % \rotatebox{90}{some rotated text} 

       \textbf{$\omega$}  & 0 & 0.9 & 0.99 & 0.999 & 1.0\\ 
%\Xhline{1.2pt}%第一条粗线
\hline
       mIoU  & 68.9 & 70.3 & 70.6 & \textbf{71.0} & 69.4\\

  % \Xhline{0.8pt}%第二条粗线
\end{tabular}
}
    \label{ema}
\end{table}

\noindent
\textbf{Qualitative ablation on the Effectiveness of Sampling Strategies.}
Figure \ref{fig:comparison} presents the results of (1) baseline, (2) baseline with instance loss \& interaction (pseudo-label regeneration) but the instances are randomly selected, and (3) DIDA, on segmenting an image containing ``confusing" entities. We observe that the naive introduction of instance loss \& interaction is helpful for common classes (\eg{}, ``fence" and ``vegetation" in red box), but lacks refinement on long-tailed/boundary pixels (the long-tailed classes of ``truck" and ``train" with some proportions of overlapping, in the yellow box), while CBS\&BPS help to distinguish between the long-tailed and overlapping categories (a more explicit segmentation of ``pole" from ``train" and ``truck" from ``train", in yellow \& gray boxes).  Please see more visualization results in our \textcolor{orange}{supplementary material}.
% will add more qualitative results in the appendix upon publication.

\noindent
\textbf{Effect of each component.}
As displayed in Table \ref{components}, when considering additional instance-level similarity discrimination, we improve the baseline \cite{DAFormer} by 0.9, revealing the effectiveness of our constructed feature bank. By applying updating strategies CBS and BPS, we explore the dominant factors of segmentation performance and witness a 0.6 gain. Finally, we introduce the pseudo-label regeneration (p.l.-reg.) to allow semantic-level and instance-level information to calibrate and balance each other, this mechanism leads to a further boost by 1.2. It's worth noticing that when applying pseudo-label regeneration with randomly selected features to update the bank (\ie, w/o CBS\&BPS), the performance gain is only 0.5. This is consistent with our previous assumption: a long-tailed distribution of instance features reduces the quality of regenerated pseudo-labels.

% \vspace{-20pt}
% \fi
% ablation 4
\renewcommand\arraystretch{1}%保证每列高度是原先的1.5倍
\begin{table}[h]\centering%星号表示双栏
\caption{\normalsize Ablation of each proposed component.}
% \vspace{3pt}
\scalebox{0.9}{
 \begin{tabular}{c c c c c|c} %设定每列的宽度以及对齐方式，并且可以做到自动换行
 
 % \rotatebox{90}{some rotated text} 
\hline
       ID  & $\mathcal{L}_{ins}$ & CBS & BPS & P.L.-Reg. & mIoU\\ 
\hline
        baseline  & $-$ & $-$ & $-$ & $-$ & 68.3\\
        I  & $\checkmark$ & $-$ & $-$ & $-$ & 69.2\\
        II  & $\checkmark$ & $\checkmark$& $-$ & $-$ & 69.6\\
        III  & $\checkmark$ & $\checkmark$  & $\checkmark$  & $-$ &  69.8\\
        IV  & $\checkmark$ & $-$  & $-$  & $\checkmark$  & 69.7\\
        V  & $\checkmark$ & $\checkmark$  & $\checkmark$  & $\checkmark$  & 71.0\\
\hline
  % \Xhline{0.8pt}%第二条粗线
\end{tabular}
}
    \label{components}
\end{table}

\section{Discussion}
\noindent\textbf{Limitation \& Future Directions.} Although DIDA has achieved state-of-the-art performance on mainstream benchmarks using a single RTX 3090, it can be inferred that incorporating larger memory support could improve overall performance (refer to Sec. \ref{bankandstrategy}). To further improve performance, one may harness recent advancements in memory technologies, which could be based on Integer Linear Programming techniques \cite{hildebrand2020autotm} or DRAM management methods \cite{peng2020capuchin}. While a detailed investigation of this aspect falls outside the scope of our current work, we firmly believe that this avenue holds significant potential for future research. In addition, another future avenue resides in the generalizability of our Dual-level Interaction approach. Considering the strong similarity among the visual understanding tasks that heavily rely on extensive annotated data (\eg{}, object detection \cite{ICCV_obj}, instance segmentation \cite{ICCV_ins_seg}, person re-identification \cite{ICCV_person}), our framework shows prominent applicability to other self-training scenarios in computer vision tasks.

\section{Conclusion}
This paper proposes a novel framework, DIDA, which for the first time introduces instance discrimination as an instance classifier to unsupervised domain adaptive semantic segmentation and interacts with semantic-level information for better noise adjustment. Unlike previous approaches, DIDA considers both semantic-level and instance-level information simultaneously. To tackle the issue of large storage consumption, we instantiate a dynamic instance feature bank and update it with carefully designed strategies. Additionally, our presented techniques of ``scattering" and ``gathering" facilitate interaction between the two levels and enable the regeneration of more robust pseudo-labels. Extensive experiments demonstrate DIDA's superiority over previous methods. DIDA obtains significant IoU gains on confusing and long-tailed classes while achieving state-of-the-art overall performance. Discovering DIDA's potential in other visual tasks is a promising direction, which will be our future work.

\section{Acknowledgment}
This research was supported in part by the National Key Research and Development Program of China under No.2021YFB3100700, the National Natural Science Foundation of China (NSFC) under Grants No. 62202340, Wuhan Knowledge Innovation Program under No. 2022010801020127.

{\small
\bibliographystyle{ieee_fullname}
\bibliography{reference.bib}
}

% \newpage

\end{document}